\documentclass[runningheads]{llncs}
\usepackage{graphicx}
\usepackage[english]{babel}
\usepackage[utf8]{inputenc}
\usepackage[caption=false,font=footnotesize]{subfig}
\usepackage[export]{adjustbox}
\usepackage{caption}
\usepackage{booktabs}
\usepackage{amsmath,amssymb}
\usepackage[pagebackref=true]{hyperref}

\usepackage[dvipsnames]{xcolor}
\usepackage{collcell}
\usepackage{hhline}
\usepackage{pgf}
\usepackage{multirow}
\usepackage{blindtext}
\usepackage{enumitem}
\usepackage{bbm}
\usepackage{siunitx}
\usepackage{arydshln}

\usepackage{accents}

\usepackage[T1]{fontenc} 
\usepackage{lipsum}

\usepackage[misc,geometry]{ifsym}
\usepackage[figuresright]{rotating}
\usepackage{booktabs}
\usepackage{svg}

\usepackage{tcolorbox}
\newtcolorbox{afancybox}[1][]{#1,colback=lightgray, colframe=black, standard jigsaw, opacityback=0.25}

\usepackage{float}
\newfloat{Code}{htbp}{loc}

\mathchardef\syphen="2D %

\newcommand{\xdownarrow}[1]{%
  {\left\downarrow\vbox to #1{}\right.\kern-\nulldelimiterspace}
}

\begin{document}

\title{Cross-Task Pretraining for Cross-Organ Cross-Scanner Adenocarcinoma Segmentation}

\titlerunning{Cross-Task PreTraining}

\author{Adrian Galdran\inst{1,2}$^\textrm{(\Letter)}$}

\authorrunning{A. Galdran}

\institute{Computer Vision Center, Universitat Autònoma de Barcelona, Spain. \and Universitat Pompeu Fabra, Barcelona, Spain. \email{adrian.galdran@upf.edu} 
}

\maketitle              %
\begin{abstract}
This short abstract describes a solution to the COSAS 2024 competition on Cross-Organ and Cross-Scanner Adenocarcinoma Segmentation from histopathological image patches. 
The main challenge in the task of segmenting this type of cancer is a noticeable domain shift encountered when changing acquisition devices (microscopes) and also when tissue comes from different organs. 
The two tasks proposed in COSAS were to train on a dataset of images from three different organs, and then predict segmentations on data from unseen organs (dataset T1), and to train on a dataset of images acquired on three different scanners and then segment images acquired with another unseen microscope.
We attempted to bridge the domain shift gap by experimenting with three different strategies: standard training for each dataset, pretraining on dataset T1 and then fine-tuning on dataset T2 (and vice-versa, a strategy we call \textit{Cross-Task Pretraining}), and training on the combination of dataset A and B. 
Our experiments showed that Cross-Task Pre-training is a more promising approach to domain generalization.

\keywords{Adenocarcinoma Segmentation \and Domain Shift}
\end{abstract}

\setcounter{footnote}{0} 

\section{Introduction}
Effective segmentation of tumoral areas from histopathology images remains a central problem in the field of computational pathology \cite{foucart_shortcomings_2023}. 
Even when circumventing the challenge of processing gigapixel-size whole-slide images by breaking them down into local patches, it is still extremely hard to train segmentation models that generalize across an array of diverse scenarios, like different acquisition devices or tissue organs. 
Such a demanding problem is known as domain shift, and has been the focus of intense research in latest years \cite{AUBREVILLE2024103155,zhou_unsupervised_2024}.

\begin{figure}[!t]
\centering
\subfloat[]{\includegraphics[width = 0.30\textwidth,valign=c]{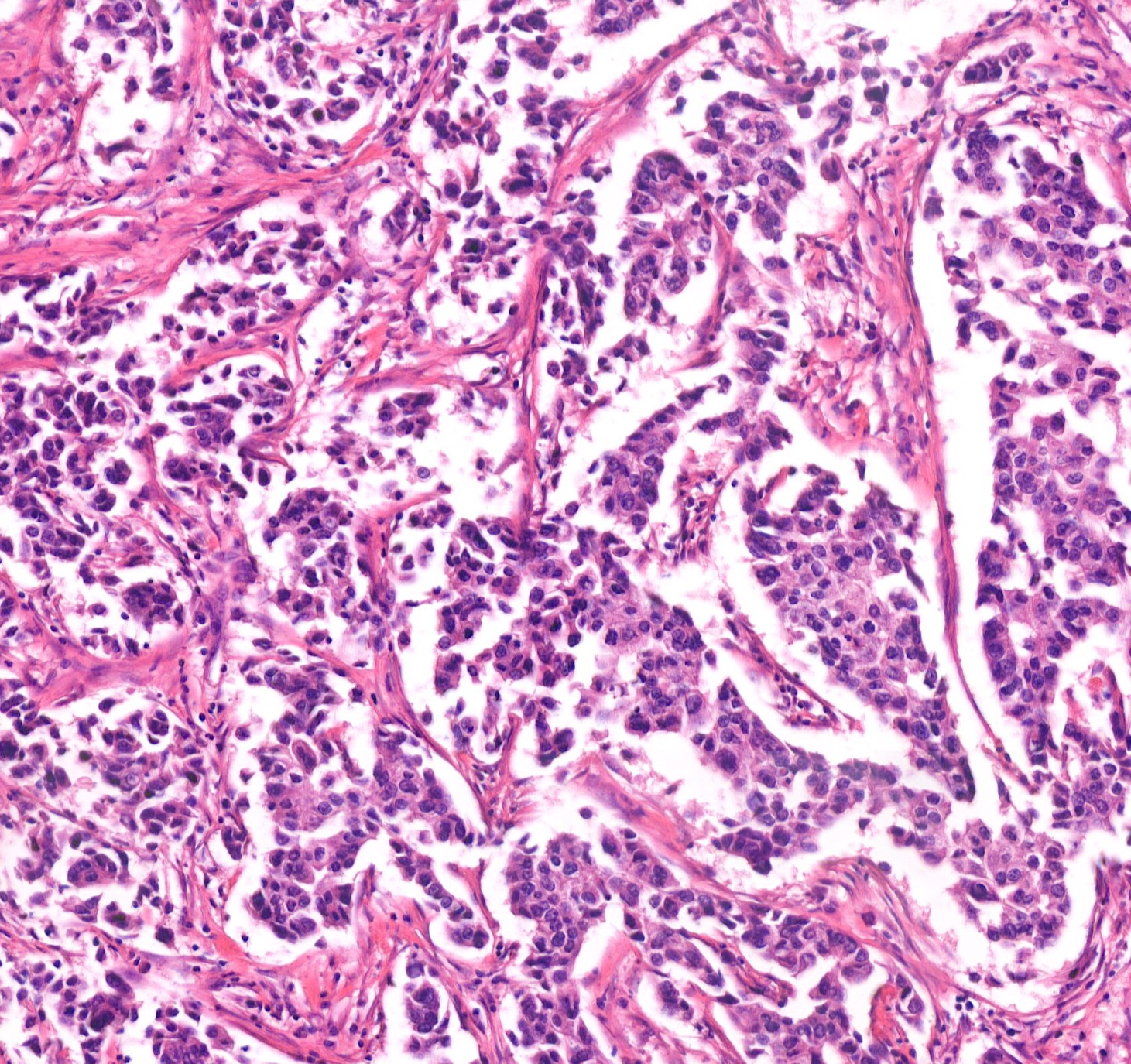}
\label{fig_o1}}
\hfil
\subfloat[]{\includegraphics[width = 0.30\textwidth,valign=c]{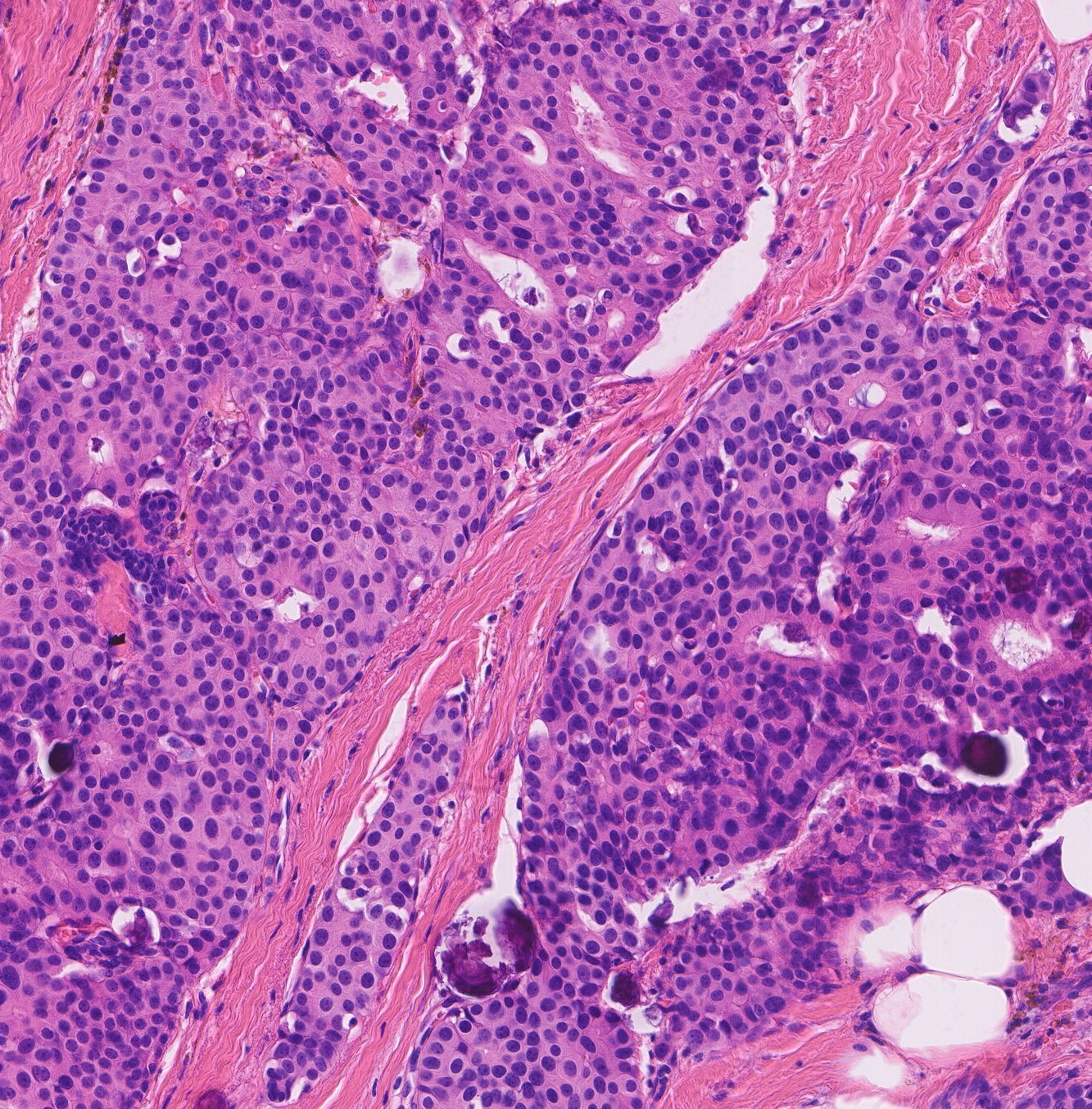}
\label{fig_o2}}
\hfil
\subfloat[]{\includegraphics[width = 0.30\textwidth,valign=c]{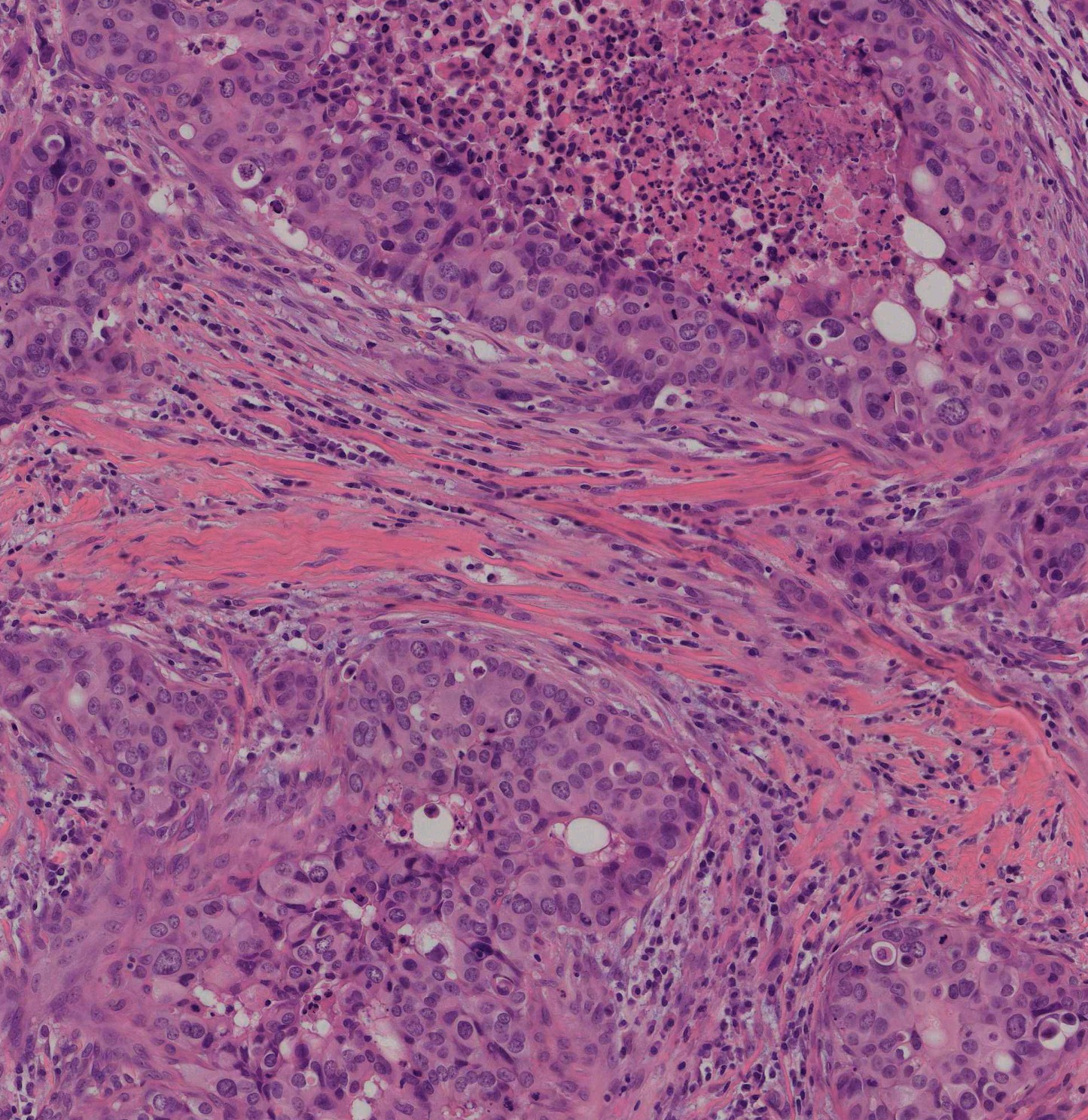}
\label{fig_o3}}
\caption{The image in (a) was digitized with a 3DHISTECH PANNORAMIC 1000 scanner, (b) was extracted from an image acquired with a KFBIO KF-PRO-400 scanner, and (c) was obtained with a TEKSQRAY SQS-600P scanner. The test set had images acquired with another three unseen scanners.
}
\label{example}
\end{figure}

In the above context, the Cross-Organ and Cross-Scanner Adenocarcinoma Segmentation (COSAS) competition was held in MICCAI 2024, in order to benchmark domain adaptation capabilities of current segmentation models. 
The challenge was divided into two tracks:
\begin{enumerate}[leftmargin=*,labelindent=0pt]
    \item \textbf{Task 1: Cross-Organ Generalization} The goal is to train a segmentation model on images corresponding two three different organs affected by adenocarcinoma (gastric adenocarcinoma, colorectal adenocarcinoma, and pancreatic ductal adenocarcinoma). The test set contains images from these three organs and another three unseen ones.
    \item \textbf{Task 2: Cross-Scanner Generalization:} In this track, the objective is to segment adenocarcinoma regions from images digitized with six different scanners. Three of them are used to generate the training set, see Fig. (\ref{example}).
\end{enumerate}

We succinctly describe next our approach to the COSAS competition, with a detailed numerical comparison of several different domain adaptation strategies.

\section{Methodology}
\subsection{Segmentation Model}
The segmentation architecture we selected for this task was a Feature-Pyramid Network with a Mix-Vision Transformer encoder pretrained on the Imagenet dataset \cite{Iakubovskii_2019}. 
Images were resized to a common resolution of $1024\times1024$, and models were all trained using an Adam optimizer with an initial learning rate of $l=1e-4$ that was cosine-annealed to zero cyclically during 30 epochs. 
The binary cross-entropy loss was minimized, as we found no improvement when adding a Dice loss component \cite{liu_we_2024}. 

\subsection{Data Engineering for improving Domain Generalization}
The above model was trained on different datasets, using three strategies, in an attempt to obtain better performance but also improve generalization ability:
\begin{enumerate}[leftmargin=*,labelindent=0pt]
\item \textbf{Standard Training}: We used the training data for each task independently. We conducted cross-validation, resulting in a five-fold ensemble submission.
\item \textbf{Cross-Task Training}: Weights after doing Standard Training for each task were used as initialization when fine-tuning a model to solve the other task.
\item \textbf{Task Union}: We simply trained for each task on the combination of both datasets. Note that validation sets were different for each task.
\end{enumerate}

\section{Experimental analysis}

\subsection{Datasets and Performance Evaluation}
The dataset for Task 1 is composed of 290 image patches showing six different adenocarcinomas. Patches have a resolution of 1500x1500 pixels, and they all come from WSIs digitised using the same scanner. Three organs are present in the training set, but the preliminary test set has 4 and the final test set six organ types, so we can expect some performance gap between cross-validation scores, our submissions to the preliminary test set (all three approaches were submitted) and performance on the final test set (only the best approach was submitted). Similarly, for Task 2 the dataset has images from six scanners, but the training set only has images from three of them, whereas the preliminary test set has four scanners and the final test set has the six of them. Performance in this case was measured by means of the standard Dice Similarity Coefficient.

\subsection{Numerical Results}
As can be seen from Table \ref{results}, a conventional strategy leads to the lowest cross-validation results in both tasks, a trend that is reproduced in the preliminary test set. 
In contrast, both using the union of the two datasets for training, and specially pretraining on one task and then fine-tuning on the other one (Cross-Task Pretraining improve performance across the board, with the latter obtaining sizeable improvements.
Note that we made allowed six submissions to the preliminary test set, but only one to the final one, so we selected the best performing approach (Cross-Task Pretraining) for our final submission.

\begin{table}[!t]
\renewcommand{\arraystretch}{1.0}
\setlength\tabcolsep{5.00pt}
 {\bf  
\begin{center}
\begin{tabular}{c c c c}
     &  \textbf{Conventional}  & \begin{tabular}{@{}c@{}}Crossed \\ Pre-Training\end{tabular} &  \begin{tabular}{@{}c@{}}Dataset \\ Union\end{tabular}  \\
\midrule
\textbf{T1: Cross-Validation}         & 75.95 $\pm$ 32.36 & 82.82 $\pm$ 23.26 & 80.57 $\pm$ 28.60         \\ 
\textbf{T1: Preliminary Set}          & 76.32     & 78.65     & 75.07          \\
\textbf{T1: Final Set}                &  n/a      & 76.07     &  n/a          \\
\midrule
\textbf{T2: Cross-Validation}         & 85.57 $\pm$ 18.34     & 86.11 $\pm$ 17.76    & 88.17 $\pm$ 16.74          \\
\textbf{T2: Preliminary Test Set}     &  81.70    & 85.69     & 85.49         \\ 
\textbf{T2: Final Test Set}           &  n/a      & 81.75     &  n/a          \\
\bottomrule
\\[-0.25cm]
\end{tabular}
\caption{Results from Cross-validating each strategy on the training set (3 organs/scanners, rows 1$^\mathrm{st}$/4$^\mathrm{th}$), submitting to the preliminary test set (4 organs/scanners, rows 2$^\mathrm{nd}$/5$^\mathrm{th}$) and to the final test set (6 organs/scanners, rows 3$^\mathrm{rd}$/6$^\mathrm{th}$).}\label{results}
\end{center}
}
\vspace{-1cm}
\label{table_1}
\end{table}

\section{Conclusions}
We summarized and experimentally tested several strategies for dealing with domain shift in adenocarcinoma segmentation from histopathological images. 
Cross-Task Pretraining, \textit{i.e.} training on one task and then fine-tuning on the other resulted in the highest performance for both tasks.

\subsection*{Acknowledgments}
A. Galdran is funded by a Ramon y Cajal fellowship RYC2022-037144-I.

\bibliographystyle{splncs04}
\bibliography{MICCAI24.bib}

\end{document}